%% file: main.tex
\documentclass[10pt,twocolumn,letterpaper]{article}

\usepackage{cvpr}              % To produce the CAMERA-READY version
\input{preamble}

\definecolor{cvprblue}{rgb}{0.21,0.49,0.74}
\usepackage[pagebackref,breaklinks,colorlinks,citecolor=cvprblue]{hyperref}

\def\paperID{*****} % *** Enter the Paper ID here
\def\confName{CVPR}
\def\confYear{2024}

\title{Window to Wall Ratio Detection using SegFormer}

\author{Zoe De Simone\\
{\tt\small zoed@mit.edu}\\
MIT\\
Cambridge, MA \\
\and
Sayandeep Biswas\\
{\tt\small biswas00@mit.edu}\\
MIT\\
Cambridge, MA \\
\and 
Oscar Wu \\
{\tt\small oscarwu@mit.edu}\\
MIT\\
Cambridge, MA \\
}

\begin{document}
\maketitle
\input{sec/0_abstract}    
\input{sec/1_intro}
\input{sec/2_methods}
\input{sec/3_results}
{
    \small
    \bibliographystyle{ieeenat_fullname}
    \bibliography{main}
}


\end{document}

%% file: preamble.tex
\usepackage[dvipsnames]{xcolor}

%% file: sec/0_abstract.tex
\begin{abstract}
Window to Wall Ratios (WWR) are key to assessing the energy, daylight and ventilation performance of buildings. Studies have shown that window area has a large impact on building performance and simulation\cite{bhandari2019influence} \cite{troup2019effect} \cite{marino2017does}. However, data to set up these environmental models and simulations is typically not available. Instead, a standard 40\% WWR is typically assumed for all buildings. 
This paper leverages existing computer vision window detection methods to predict WWR of buildings from external street view images using semantic segmentation, demonstrating the potential for adapting established computer vision technique in architectural applications.
\end{abstract}

%% file: sec/1_intro.tex
\section{Introduction}
\label{sec:intro}
Windows play a significant role in the performance and comfort of buildings, transmitting daylight, heat, and wind. However, information about window placement and window-to-wall ratio (WWR), which are necessary to set up building performance simulation models (such as daylight, energy, and thermal comfort models), is typically not publicly available, and instead, rules of thumb of 40\% WWR are applied to all buildings, regardless of location and climate \cite{szczesniak2022method}. To improve the accuracy and data availability of building characteristics, previous research has centered around façade geometry extraction using image processing techniques or specialized equipment such as 3D laser scanning \cite{fruh2003constructing, fruh2004automated, marston2015testing}, which are time intensive and require dedicated hardware.

Computer vision methods to process façade images to obtain WWR benefit from scalability and availability of publicly labeled datasets of façade images such as TSG-2002, TSG-6003, ZuBuDO4, CMP, and ECP \cite{isola2017image, li2020window}, which can be augmented for increased accuracy in specific climates, through private manually labeled images captured from Google Street View \cite{tarkhan2022capturing}. However, the time-consuming nature of manually labeling building façades limits the number of high-quality labeled data for WWR predictions. Fortunately, several foundation models trained on much larger datasets such as ImageNet, ADE20K \cite{zhou2017scene}, and more can be leveraged to address this data scarcity. 

Image-based approaches revolve mainly around 3 types: 1) semantic parsing of images with regular structure using Conditional random fields (CRF)\cite{tylevcek2013spatial}, 2) parsing and classification of  image pixels based on their function such as sky, wall, door, and roof, window \cite{cohen2014efficient}, and 3) Convolutional Neural Network (CNN) models using heatmaps to detect window corners and center points to then form bounding boxes surrounding each window \cite{li2020window, tarkhan2022capturing}. There has been significant progress in the field, in terms of model architectures and datasets leading to better accuracies. However, there is still room for further improvement as results from the recent paper by Tarkhan et al. have only 72\% of buildings detected within the 10\% error range \cite{tarkhan2022capturing}. Furthermore, certain subsets of building façades such as complete glass finishes (WWR greater than 70-90\%) have significantly worse predictions. 

Another major obstacle spanning WWR approaches is related to the calculation of the window and wall area. Generally, photos of buildings are taken at street level and at an angle. In order to precisely calculate the WWR, and accurately weigh each portion of the facade equally, images then have to be re-projected such that they are frontal facing.  Previous work by Tarkhan et al. handled orthogonal transformations by computing edges, principal vanishing points, computing homography, and warping the images to fronto-parallel view with orthogonal axes. Lastly, images were cropped to the facade area of the building, such that the image area is a proxy for the facade area \cite{tarkhan2022capturing}. This methodology, however, presents its own challenges, only working on images that are heavily skewed.

To address the aforementioned challenges of window and facade detection, the efforts of this paper are three-fold: 1) we create an augmented semantic segmentation dataset with window and facade labels, 2) we train several model architectures, including Fully Convolutional Networks (FCNs) and SegFormer to detect windows and facade areas, 3) we develop a corner point method to reproject images of buildings to fronto-parallel views, which generalizes to all skewed images. 

While we did not develop novel ML architectures in this study, we demonstrate the potential for adapting established computer vision techniques to address research questions in architecture, an area where these methods have thus far been underutilized. We note that the methods and results presented here are part of an early stage research project and should be considered preliminary. 
We share our opensource code in a github repository: \href{https://github.com/zoedesimone/wwr-semantic-segmentation}{https://github.com/zoedesimone/wwr-semantic-segmentation}.

%% file: sec/2_methods.tex
\section{Introduction}
\label{sec:formatting}
\subsection{Dataset Preparation}
The public datasets mentioned previously were compiled to create a combined dataset of 2806 façade images. The raw dataset for each façade consists of a JPEG and a JSON file. The JSON file contains the coordinates of the polygons used to label the windows. This format is not amenable to the semantic segmentation models which require a JPEG input and a PNG image as the label. Therefore, the first step involved parsing the JSON files for each image to generate the labels shown in Figure \ref{fig:label_me}. 

\subsubsection{FCN}
\begin{figure}
    \centering
    \includegraphics[width=0.4\textwidth, height=4cm]{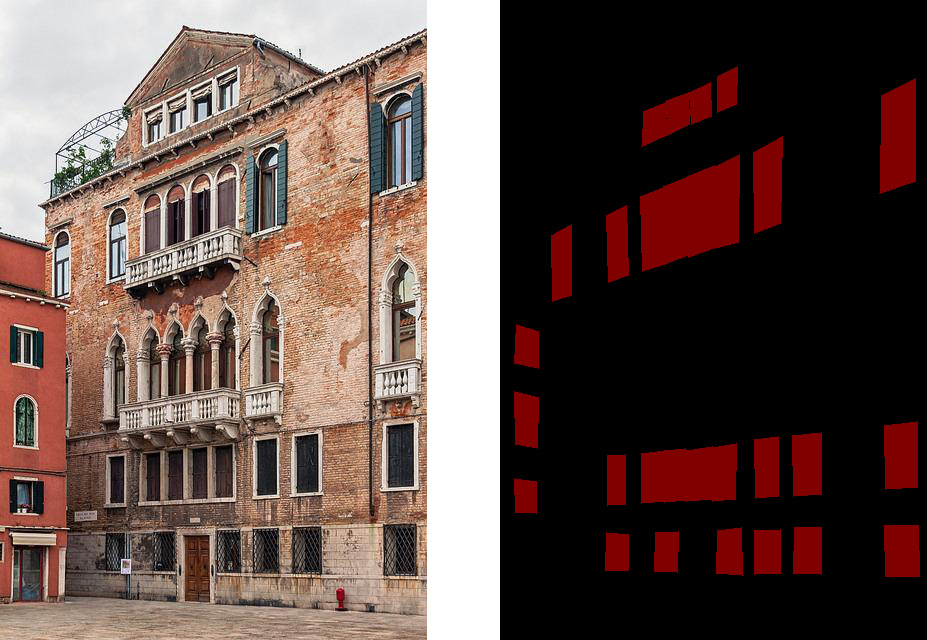}
   \caption{Dataset preparation: JPEG of building façade (left) and PNG with marked windows(right).}
    \label{fig:label_me}
\end{figure}

The next step for dataset preparation for the FCN architecture is to resize and normalize the input images. The input size was chosen to be 520-by-520, followed by a re-scaling to ensure pixel values were between 0 and 1. Next, the RGB channels were normalized using the following parameters: $[\mu_R, \sigma_R]  = [0.485, 0.229], [\mu_G, \sigma_G]  = [0.456, 0.224], \text{and } [\mu_B, \sigma_B]  = [0.406, 0.225]$. The parameters for resizing, scaling, and normalizing were based on the dataset used to pre-train the FCN models. The labels for FCN were obtained from the labeled PNGs as shown in Figure \ref{fig:label_me}. Simple logic was used to build a mask where 1 represented window pixels (red) and 0 represented the background (black). Next, the obtained mask was resized to also be 520-by-520.

\subsubsection{SegFormer}
The next step in data preparation for the SegFormer architecture is to generate the labeled dataset containing pixel-by-pixel encodings from the predefined ADE20K classes \cite{zhou2017scene}, including exterior windows and building perimeters. The same normalized images as in the FCN were used.

\begin{figure}
\centering
    \includegraphics[width=0.4\textwidth, height=4cm]{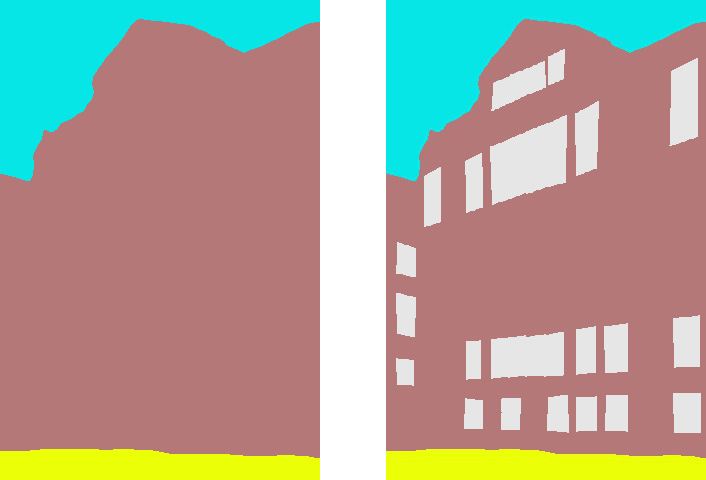}
    \caption{Enhancing ADE20K labels by superimposing using \textit{SegFormer} predicted labels (Left) onto our labeled windows (Right).}
\label{fig:superimp}
\end{figure}

The ADE20K dataset contained labels of building perimeters and windows, however, only interior windows were labeled and exterior windows are completely excluded. We obtain SegFormer model labels including exterior windows, by using a pre-trained SegFormer with a ResNet50dilated + PPMdeepsup encoder and decoder architecture to detect the building area in the images, and overlaying window masks developed for the FCN architecture as shown in Figure \ref{fig:superimp}. Finally, we converted the output to greyscale masks encoding a single number per pixel, representing the encoded semantic label in the ADE20K semantic categories.

Lastly, we randomly crop, pad and resize both image and segmentation maps to the same size 512-by-512.

\subsection{Dataset Distributions}
\begin{figure}
    \includegraphics[width=.4\textwidth, height=6cm]{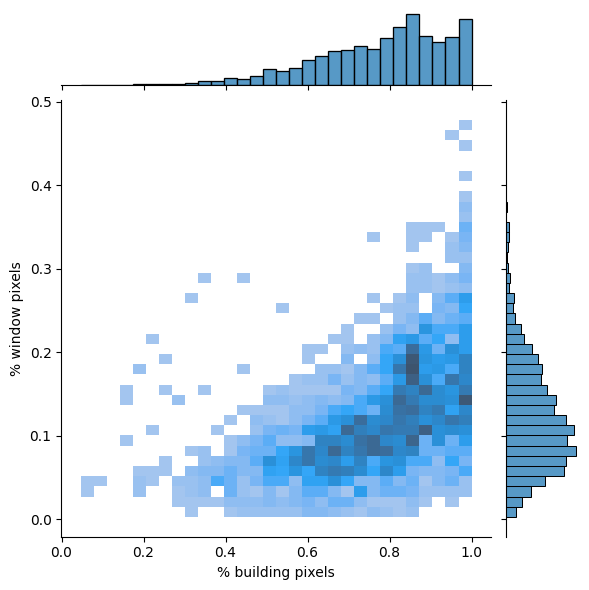}
    \caption{Window pixels \% vs Building Pixel \%}
    \label{fig:pixelperc}
\end{figure}

Analyzing our dataset, We can see that the mean WWR is 0.158 and mode WWR is 0.13, with nearly all of the images below 40\% WWR, confirming our hypothesis that existing 40\% WWR rules of thumb are not error-free.
From the distribution of the percentage of window and wall pixels in each image as shown in Figure \ref{fig:pixelperc}, we can see that photos are not cropped to the building's façade. Instead, photos contain building surroundings, and therefore the image area should not be used as a proxy for the façade area. 

For each image, we compute the WWR by calculating the façade area \(A_f\) and window area \(A_w\):
\begin{equation*}
    WWR = \frac{A_w}{A_f+A_w}
\end{equation*}
The façade and window area are computed by counting the number of {pink} and {grey} pixels in the image.

\subsection{Model and Training}
In this work, we showcase how these models can be trained with limited resources, within the Google Colab computational constraints. We utilized a randomly selected 10\% subset consisting of 280 images to train our models. This subset was randomly divided into a training, validation, and testing set using an 80:10:10 split. After each epoch, the Intersection over Union (IoU) metric is computed on the validation set. After completion of training, the epoch with the highest IoU score on the validation set is chosen as the final model. Further details specific to each architecture are discussed below.

\begin{equation*}
    IoU = \frac{\text{Area of Overlap}}{\text{Area of Union}}
\end{equation*}

\subsubsection{FCN}
The FCN models used in this work were imported from the torchvision library and were developed by \citeauthor{long2015fully}. The models were pre-trained on a subset of COCO, using only the 20 categories that are present in the Pascal VOC dataset. For our application, we have a single-channel output for the windows compared to the pre-trained model that has 21 channels including the background. Therefore, the final convolution layer was modified to have the correct dimensions. The outputs of these models were then passed through a sigmoid layer to guarantee values between 0 and 1. The Adam optimizer with a binary cross-entropy loss was used to train the model.

\subsubsection{SegFormer}
SegFormer consists of a hierarchical Transformer encoder and a lightweight all-MLP decode head. The SegFormer models used in this work were imported from HuggingFace and were developed by \citeauthor{xie2021segformer}. The code was inspired from \citeauthor{Rogge_Transformers_Tutorials_2020} developed using PyTorch by Rogge and Niels. We load the encoder of the model, namely b0, b1, and b5 \cite{xie2021segformer}, with weights pre-trained on ImageNet-1k, and fine-tune it together with the decoder head, which starts with randomly initialized weights. The model was finetuned using the AdamW optimizer. The learning rate of 0.00006 reported in the SegFormer paper was used. Cross entropy loss was used for training and the window IoU was used to compute the validation and test set performance. 

%% file: sec/3_results.tex
\section{Results}
\subsection{Semantic Segmentation}
Table \ref{model-table} shows a table summarizing our models' performance. A preliminary study was conducted without comprehensive hyperparameter optimization, primarily using default settings. The FCN model with a ResNet-50 backbone achieved the best performance, recording an IoU score of 0.66 on the test set with a 0.0005 learning rate.

\begin{table}[t]
\caption{Summary of training details and performance of various FCN and SegFormer(Seg) models varying model choice, pretraining, number of epochs, IoU on testing Data and Learning Rate (LR).}
\label{model-table}
\vskip 0.15in
\begin{center}
\begin{small}
\begin{sc}
\begin{tabular}{lcccr}
\toprule
Model&Pretrain&Epochs&IoU Test&LR\\
\midrule
Seg-b0 & $\surd$& 10& 0.58& 0.00006 \\
Seg-b1 & $\surd$& 10 & 0.62& 0.00006\\
Seg-b5 & $\surd$   & 10 & 0.65& 0.00006  \\
Resnet-50    &$\surd$ & 10 & 0.61& 0.005        \\
Resnet-50    &$\surd$  & 100 & 0.66& 0.0005\\
Resnet-50     &$\times$  & 100& 0.60& 0.005 \\
Resnet-101    &$\surd$  & 100& 0.59& 0.005  \\
\bottomrule
\end{tabular}
\end{sc}
\end{small}
\end{center}
\vskip -0.1in
\end{table}

Table \ref{model-table} details the training and performance metrics of various models, indicating the significance of model size for SegFormer, where a larger model (b5) outperformed others. Conversely, for FCN models, a ResNet-50 backbone was more effective than the larger ResNet-101. Learning rate emerged as a crucial factor, with a lower rate significantly enhancing the FCN ResNet-50 model's IoU score by nearly 6\%. The impact of pre-training was minimal, suggesting that other hyperparameters and model choices might offer substantial performance gains with dedicated optimization.

\begin{figure}
    \centering
    \includegraphics[width=0.48\textwidth, height=5cm]{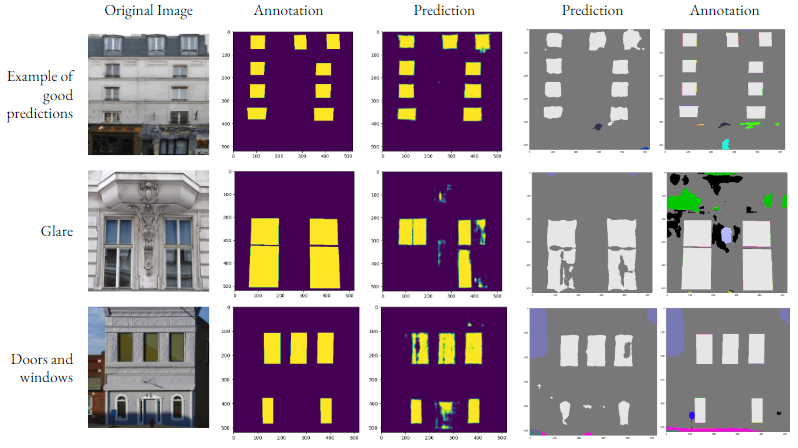}
    \caption{Prediction generated by FCN (left) and SegFormer (right) models on images in the test set}
    \label{fig:visual_results}
\end{figure}

Figure \ref{fig:visual_results} displays segmented images from our models, highlighting failure modes such as misclassification due to glare on windows, doors mistaken for windows, and issues with visible interiors and occlusions. These issues underscore the potential benefits of larger datasets and data augmentation. Our approach used a single-fold, which serves for a proof-of-concept; however, cross-fold training and extrapolative data splits are recommended for improved results and universal applicability.

\subsection{Robustness Assessment}
\begin{figure}
    \centering
    \includegraphics[width=0.4\textwidth, height=4cm]{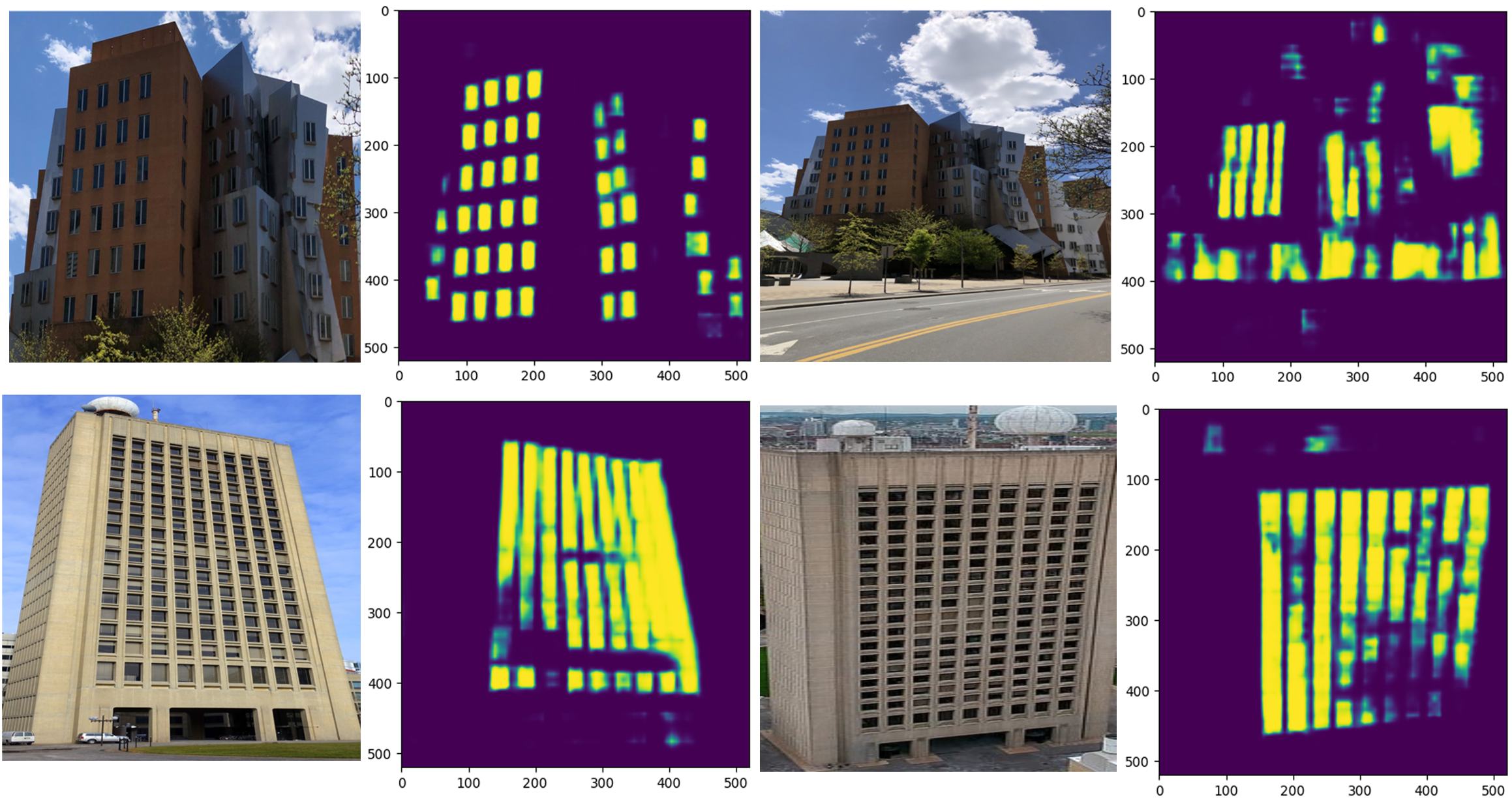}
    \caption{FCN ResNet-50 predictions on varying levels of zoom and lighting conditions.}
    \label{fig:stata_building}
\end{figure}

We evaluated the robustness of our top FCN ResNet-50 model under different times of day, zoom levels, and lighting conditions (Figure \ref{fig:stata_building}). The model shows good resilience to lighting variations but struggles with varying zoom levels. Specifically, it performs better on close-ups of building facades, reflecting the training data's composition of 80\% building pixels (Figure \ref{fig:pixelperc}). To enhance zoom level robustness, incorporating more zoomed-out images or generating synthetic images with artificial backgrounds into the training set could be beneficial.

\subsection{Perspective Correction}
\begin{figure}
    \centering
    \includegraphics[width=0.4\textwidth, height=3cm]{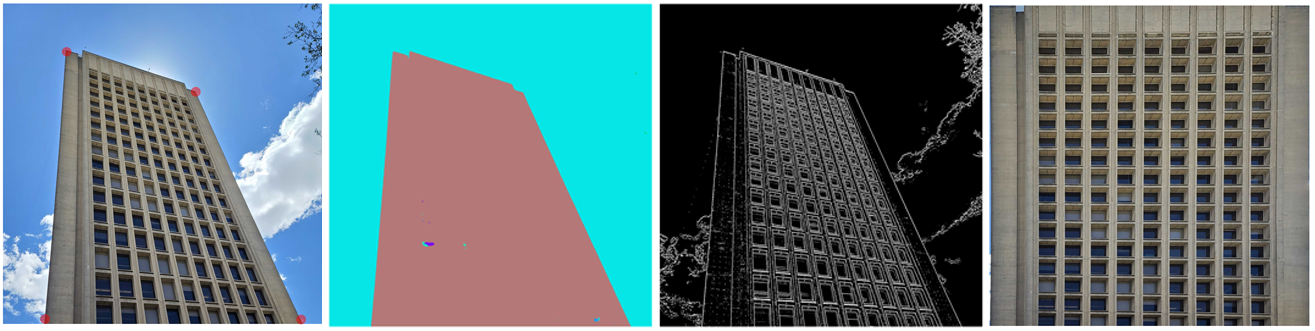}
    \caption{1) Original image with building corners identified for the four point perspective transform algorithm; 2)Semantic segmentation mask and 3) Edge detection to identify building corners for perspective correction. 4) Building image after perspective correction.}
    \label{fig:perspec_1}
\end{figure}

For accurate WWR calculations, correct estimation of window and wall areas is essential. Street-level images often suffer from perspective distortion due to oblique shooting angles, making parts of the facade closer to the camera appear larger and distorting the true proportions. We address this by reprojecting images to a fronto-parallel view using a four-point perspective transformation in OpenCV, exemplified by adjusting an angled image of a building (Figure \ref{fig:perspec_1}). This method simulates a perpendicular view of the facade, equalizing distances from the observer and correcting the proportions for more precise WWR estimations. However, slight distortions remain, potentially affecting WWR accuracy by compressing windows at the building's top more than those at the bottom.

Additionally, we investigated automatic building corner detection for the transformation algorithm, exploring semantic segmentation and edge detection techniques (Figure \ref{fig:perspec_1}). While segmentation models are good for identifying the bulk of objects, as highlighted by Zou et al. \cite{zou_end--end_2022}, they may not accurately delineate precise boundaries for corner detection. Edge detection can provide more exact corner locations but is sensitive to image noise. Therefore, further exploration is warranted to investigate the potential benefits of combining both semantic segmentation and edge detection methodologies to enhance corner detection performance.

\section{Future Work}
Improving our models' performance hinges on addressing training dataset limitations, such as missing or inaccurately labeled windows. Future strategies should include data cleaning for a refined dataset, employing k-fold cross-validation to better evaluate model robustness across dataset subsets, and multi-fidelity training, which leverages a tiered approach by initially training on a broader, less precise dataset before fine-tuning on a more accurately labeled subset.

\section{Conclusion}
Prior approaches to Window to Wall Ratio (WWR) detection have primarily targeted window identification, using perspective correction and cropping to estimate wall areas—a method prone to inaccuracies from including non-wall elements. Our research introduces a semantic segmentation dataset with detailed window labels and enhanced with ADE20K semantic categories, aiming to accurately segment both windows and walls. We trained FCN and SegFormer models on this dataset, achieving more precise window and wall detection. Despite their effectiveness, challenges remain, including the misclassification of reflective windows and doors as windows, underscoring the need for ongoing refinement. We note that the methods and results presented here are part of an early stage research project and should be considered preliminary.

\section{Acknowledgements}
We thank Nada Tarkhan for invaluable discussions on window-to-wall ratio and urban extraction methods. This material is based upon work supported by the National Science Foundation Graduate Research Fellowship under Grant No. 2141064. Any opinion, findings, and conclusions or recommendations expressed in this material are those of the authors(s) and do not necessarily reflect the views of the National Science Foundation.

%% file: main.bbl
\begin{thebibliography}{17}
\providecommand{\natexlab}[1]{#1}
\providecommand{\url}[1]{\texttt{#1}}
\expandafter\ifx\csname urlstyle\endcsname\relax
  \providecommand{\doi}[1]{doi: #1}\else
  \providecommand{\doi}{doi: \begingroup \urlstyle{rm}\Url}\fi

\bibitem[Bhandari and Tadepalli(2019)]{bhandari2019influence}
Nikhil Bhandari and Srinivas Tadepalli.
\newblock Influence of window-to-wall ratio on building energy load with daylight utilization for west facade in office building in hot and dry climate of india: A simulation-based approach.
\newblock In \emph{In: Proceedings of the International Conference of Architectural Science Association}, pages 615--624, 2019.

\bibitem[Cohen et~al.(2014)Cohen, Schwing, and Pollefeys]{cohen2014efficient}
Andrea Cohen, Alexander~G Schwing, and Marc Pollefeys.
\newblock Efficient structured parsing of facades using dynamic programming.
\newblock In \emph{Proceedings of the IEEE Conference on Computer Vision and Pattern Recognition}, pages 3206--3213, 2014.

\bibitem[Fr{\"u}h and Zakhor(2003)]{fruh2003constructing}
Christian Fr{\"u}h and Avideh Zakhor.
\newblock Constructing 3d city models by merging aerial and ground views.
\newblock \emph{IEEE computer graphics and applications}, 23\penalty0 (6):\penalty0 52--61, 2003.

\bibitem[Fr{\"u}h and Zakhor(2004)]{fruh2004automated}
Christian Fr{\"u}h and Avideh Zakhor.
\newblock An automated method for large-scale, ground-based city model acquisition.
\newblock \emph{International Journal of Computer Vision}, 60:\penalty0 5--24, 2004.

\bibitem[Isola et~al.(2017)Isola, Zhu, Zhou, and Efros]{isola2017image}
Phillip Isola, Jun-Yan Zhu, Tinghui Zhou, and Alexei~A Efros.
\newblock Image-to-image translation with conditional adversarial networks.
\newblock In \emph{Proceedings of the IEEE conference on computer vision and pattern recognition}, pages 1125--1134, 2017.

\bibitem[Li et~al.(2020)Li, Zhang, Liu, Zhang, Zou, and Fang]{li2020window}
Chuan-Kang Li, Hong-Xin Zhang, Jia-Xin Liu, Yuan-Qing Zhang, Shan-Chen Zou, and Yu-Tong Fang.
\newblock Window detection in facades using heatmap fusion.
\newblock \emph{Journal of Computer Science and Technology}, 35:\penalty0 900--912, 2020.

\bibitem[Long et~al.(2015)Long, Shelhamer, and Darrell]{long2015fully}
Jonathan Long, Evan Shelhamer, and Trevor Darrell.
\newblock Fully convolutional networks for semantic segmentation.
\newblock In \emph{Proceedings of the IEEE conference on computer vision and pattern recognition}, pages 3431--3440, 2015.

\bibitem[Marino et~al.(2017)Marino, Nucara, and Pietrafesa]{marino2017does}
C Marino, A Nucara, and M Pietrafesa.
\newblock Does window-to-wall ratio have a significant effect on the energy consumption of buildings? a parametric analysis in italian climate conditions.
\newblock \emph{Journal of Building Engineering}, 13:\penalty0 169--183, 2017.

\bibitem[Marston et~al.(2015)Marston, Turner, Zakhor, Baumann, and Haves]{marston2015testing}
Annie Marston, Eric Turner, Avideh Zakhor, Oliver Baumann, and Philip Haves.
\newblock \emph{Testing rapmod: Can a portable scanner collect exisitng building data and create an energy model faster and more accurately than a human}.
\newblock eScholarship, University of California, 2015.

\bibitem[Rogge(2020)]{Rogge_Transformers_Tutorials_2020}
Niels Rogge.
\newblock {Transformers Tutorials"}, 2020.

\bibitem[Szcze{\'s}niak et~al.(2022)Szcze{\'s}niak, Ang, Letellier-Duchesne, and Reinhart]{szczesniak2022method}
Jakub~T Szcze{\'s}niak, Yu~Qian Ang, Samuel Letellier-Duchesne, and Christoph~F Reinhart.
\newblock A method for using street view imagery to auto-extract window-to-wall ratios and its relevance for urban-level daylighting and energy simulations.
\newblock \emph{Building and Environment}, 207:\penalty0 108108, 2022.

\bibitem[Tarkhan et~al.(2022)Tarkhan, Letellier-Duchesne, and Reinhart]{tarkhan2022capturing}
Nada Tarkhan, Samuel Letellier-Duchesne, and Christoph Reinhart.
\newblock Capturing fa{\c{c}}ade diversity in urban settings using an automated window to wall ratio extraction and detection workflow.
\newblock In \emph{2022 Annual Modeling and Simulation Conference (ANNSIM)}, pages 706--717. IEEE, 2022.

\bibitem[Troup et~al.(2019)Troup, Phillips, Eckelman, and Fannon]{troup2019effect}
Luke Troup, Robert Phillips, Matthew~J Eckelman, and David Fannon.
\newblock Effect of window-to-wall ratio on measured energy consumption in us office buildings.
\newblock \emph{Energy and Buildings}, 203:\penalty0 109434, 2019.

\bibitem[Tyle{\v{c}}ek and {\v{S}}{\'a}ra(2013)]{tylevcek2013spatial}
Radim Tyle{\v{c}}ek and Radim {\v{S}}{\'a}ra.
\newblock Spatial pattern templates for recognition of objects with regular structure.
\newblock In \emph{Pattern Recognition: 35th German Conference, GCPR 2013, Saarbr{\"u}cken, Germany, September 3-6, 2013. Proceedings 35}, pages 364--374. Springer, 2013.

\bibitem[Xie et~al.(2021)Xie, Wang, Yu, Anandkumar, Alvarez, and Luo]{xie2021segformer}
Enze Xie, Wenhai Wang, Zhiding Yu, Anima Anandkumar, Jose~M Alvarez, and Ping Luo.
\newblock Segformer: Simple and efficient design for semantic segmentation with transformers.
\newblock \emph{Advances in Neural Information Processing Systems}, 34:\penalty0 12077--12090, 2021.

\bibitem[Zhou et~al.(2017)Zhou, Zhao, Puig, Fidler, Barriuso, and Torralba]{zhou2017scene}
Bolei Zhou, Hang Zhao, Xavier Puig, Sanja Fidler, Adela Barriuso, and Antonio Torralba.
\newblock Scene parsing through ade20k dataset.
\newblock In \emph{Proceedings of the IEEE conference on computer vision and pattern recognition}, pages 633--641, 2017.

\bibitem[Zou et~al.(2022)Zou, Liu, and Lee]{zou_end--end_2022}
Xueyan Zou, Haotian Liu, and Yong~Jae Lee.
\newblock End-to-{End} {Instance} {Edge} {Detection}, 2022.
\newblock arXiv:2204.02898 [cs] version: 1.

\end{thebibliography}
